\documentclass[lettersize,journal]{IEEEtran}

\usepackage{CJKutf8}
\usepackage{amsmath}
\usepackage{amssymb}
\usepackage{algorithm}
\usepackage{enumitem}
\usepackage{picinpar}
\usepackage[noend]{algpseudocode}
\usepackage{graphicx}
\usepackage{caption}

\usepackage{algorithmicx}
\usepackage{subfigure}
\usepackage{multirow}
\usepackage{color}
\usepackage{xcolor}
\usepackage{balance}
\usepackage{enumitem}
\usepackage{hhline}
\usepackage[normalem]{ulem}
\usepackage{booktabs}
\usepackage{wrapfig}
\usepackage{cancel}
\usepackage{hyperref}
\usepackage{makecell}
\usepackage[switch,columnwise]{lineno}

\newcommand{\bL}{\ensuremath{\mathcal{L}}}

\newcommand{\bG}{\ensuremath{\mathcal{G}}}
\newcommand{\bN}{\ensuremath{\mathcal{N}}}

\newcommand{\bT}{\ensuremath{\mathcal{T}}}

\newcommand{\bP}{\ensuremath{\mathcal{P}}}

\newcommand{\bC}{\ensuremath{\mathcal{C}}}
\newcommand{\bD}{\ensuremath{\mathcal{D}}}

\renewcommand{\vec}[1]{\ensuremath{\mathbf{#1}}}

\newcommand{\stitle}[1]{\vspace{1mm} \noindent {\bf #1}}

\newcommand{\eg}{{\it e.g.}}

\newcommand{\ie}{{\it i.e.}}

\newcommand{\method}[1]{\textsc{#1}}
\newcommand{\model}{\method{GraphPrompt}{}}
\newcommand{\nmodel}{\method{GraphPrompt+}{}}

\newcommand{\eat}[1]{}

\newcommand{\stkout}[1]{\ifmmode\text{\sout{\ensuremath{#1}}}\else\sout{#1}\fi}

\AtBeginDocument{%
  }

\begin{document}

\title{Generalized Graph Prompt:\\ Toward a Unification of Pre-Training and Downstream Tasks on Graphs}

\author{Xingtong~Yu,
        Zhenghao~Liu,
        Yuan~Fang,~\IEEEmembership{Senior Member,~IEEE,}
        Zemin~Liu,
        Sihong~Chen,
        and Xinming~Zhang,~\IEEEmembership{Senior Member,~IEEE,}
\IEEEcompsocitemizethanks{
\IEEEcompsocthanksitem Xingtong Yu is with the University of Science and Technology of China, Hefei, Anhui 230052, China, and also with Singapore Management University, Singapore 188065 (email: yxt95@mail.ustc.edu.cn). Work was done as a visiting student at Singapore Management University. 
\IEEEcompsocthanksitem Zhenghao Liu is with the University of Science and Technology of China, Hefei, Anhui 230052, China (email: salzh@mail.ustc.edu.cn). 
\IEEEcompsocthanksitem Yuan Fang is with Singapore Management University, Singapore 188065 (email: yfang@smu.edu.sg).
\IEEEcompsocthanksitem Zemin Liu is with the National University of Singapore, Singapore 119077 (e-mail: liu.zemin@hotmail.com).
\IEEEcompsocthanksitem Sihong Chen is with the Tecent AI, China, Shenzhen, Guangdong 518063 (email: cshwhale@sina.com).
\IEEEcompsocthanksitem Xinming Zhang is with the University of Science and Technology of China, Hefei, Anhui 230052, China (email: xinming@ustc.edu.cn).
}
\thanks{Corresponding author: Yuan Fang; Xinming Zhang.}
\thanks{Manuscript received April 19, 2021; revised August 16, 2021.}}

\markboth{IEEE TRANSACTIONS ON KNOWLEDGE AND DATA ENGINEERING}%
{Shell \MakeLowercase{\textit{et al.}}: A Sample Article Using IEEEtran.cls for IEEE Journals}

\IEEEtitleabstractindextext{%
\begin{abstract}
Graphs can model complex relationships between objects, enabling a myriad of Web applications such as online page/article classification and social recommendation. 
While graph neural networks (GNNs) have emerged as a powerful tool for graph representation learning, in an end-to-end supervised setting, their performance heavily relies on a large amount of task-specific supervision.
To reduce labeling requirement, the ``pre-train, fine-tune'' and ``pre-train, prompt'' paradigms have become increasingly common. In particular, prompting is a popular alternative to fine-tuning in natural language processing, which is designed to narrow the gap between pre-training and downstream objectives in a task-specific manner. However, existing study of prompting on graphs is still limited, lacking a universal treatment to appeal to different downstream tasks. In this paper, we propose \model, a novel pre-training and prompting framework on graphs. \model\ not only unifies pre-training and downstream tasks into a common task template, but also employs a learnable prompt to assist a downstream task in locating the most relevant knowledge from the pre-trained model in a task-specific manner. 
In particular, \model\ adopts simple yet effective designs in both pre-training and prompt tuning: During pre-training, a link prediction-based task is used to materialize the task template; during prompt tuning, a learnable prompt vector is applied to the \textsc{ReadOut} layer of the graph encoder.  
To further enhance \model\ in these two stages, we extend it into \nmodel\ with two major enhancements.
First, we generalize a few popular graph pre-training tasks beyond simple link prediction to broaden the compatibility with our task template. Second, we propose a more generalized prompt design that incorporates a series of prompt vectors within every layer of the pre-trained graph encoder, in order to capitalize on the hierarchical information across different layers beyond just the readout layer.  
Finally, we conduct extensive experiments on five public datasets to evaluate and analyze \model\
and \nmodel.
\end{abstract}
}
\maketitle

\IEEEdisplaynontitleabstractindextext

%
\IEEEpeerreviewmaketitle

\section{Introduction}\label{sec:intro}
The ubiquitous Web is becoming the ultimate data repository, capable of linking a broad spectrum of objects to form gigantic and complex graphs \cite{wang2023brave}. The prevalence of graph data enables a series of downstream tasks for Web applications, ranging from online page/article classification to friend recommendation in social networks.
Modern approaches for graph analysis generally resort to graph representation learning including graph embedding and graph neural networks (GNNs). Earlier graph embedding approaches \cite{perozzi2014deepwalk,tang2015line,grover2016node2vec} usually embed nodes on the graph into a low-dimensional space, in which the structural information such as the proximity between nodes can be captured \cite{cai2018comprehensive,bo2023survey}.
More recently, GNNs \cite{kipf2016semi,hamilton2017inductive,yu2023learning,zhang2023dropconn,10109130,wang2022searching,fang2023STWave} have emerged as the state of the art for graph representation learning. Their key idea boils down to a message-passing framework, in which each node derives its representation by receiving and aggregating messages from its neighboring nodes recursively \cite{wu2020comprehensive,DBLP:conf/nips/FangLGLZW023,Mo_ICML_2023,zhang2024two,fang2023spatio,wu2022state}. 

\stitle{Graph pre-training.}
Typically, GNNs work in an end-to-end manner, and their performance depends heavily on the availability of large-scale, task-specific labeled data as supervision. This supervised paradigm presents two problems. First, task-specific supervision is often difficult or costly to obtain. Second, to deal with a new task, the weights of GNN models need to be retrained from scratch, even if the task is on the same graph.  
To address these issues, pre-training GNNs \cite{hu2020gpt,Hu2020Strategies,qiu2020gcc,lu2021learning,xu2022ccgl,liu2021self} has become increasingly popular, inspired by pre-training techniques in language and vision applications \cite{dong2019unified,bao2022beit,hu2023survey}. 
The pre-training of GNNs leverages self-supervised learning on more readily available label-free graphs (\ie, graphs without task-specific labels), and learns intrinsic graph properties that intend to be general across tasks and graphs in a domain. In other words, the pre-training extracts a task-agnostic prior, and can be used to initialize model weights for a new task. Subsequently, the initial weights can be quickly updated through a lightweight fine-tuning step on a smaller number of task-specific labels.

However, the ``pre-train, fine-tune'' paradigm suffers from the problem of inconsistent objectives between  pre-training and downstream tasks, resulting in suboptimal performance \cite{liu2021pre}. 
On one hand, the pre-training step aims to preserve various intrinsic graph properties such as node/edge features \cite{Hu2020Strategies,hu2020gpt}, node connectivity/links \cite{hamilton2017inductive,hu2020gpt,lu2021learning}, and local/global patterns \cite{qiu2020gcc,Hu2020Strategies,lu2021learning}.
On the other hand, the fine-tuning step aims to reduce the task loss, \ie, to fit the ground truth of the downstream task.
To narrow the gap between pre-training and downstream tasks, prompting \cite{brown2020language} has first been proposed for language models, which is a natural language instruction designed for a specific downstream task to elicit the semantic relevance between the task and the language model \cite{zhu2022prompt}. 

\stitle{Research problem and challenges.}
To address the divergence between graph pre-training and various downstream tasks, in this paper we investigate the design of pre-training and prompting for graph neural networks. We aim for a unified design that can suit different downstream tasks flexibly.  
This problem is non-trivial due to the following two challenges.

Firstly, to enable effective knowledge transfer from the pre-training to a downstream task, it is desirable that the pre-training step preserves graph properties that are compatible with the given task. However, since different downstream tasks often have different objectives, \emph{how do we unify pre-training with various downstream tasks on graphs}, so that a single pre-trained model can universally support different tasks? That is, we try to convert the pre-training task and downstream tasks to follow the same ``template''. Using pre-trained language models as an analogy, both their pre-training and downstream tasks can be formulated as masked language modeling. 

Secondly, under the unification framework, it is still important to identify the distinction between different downstream tasks, in order to attain task-specific optima. For pre-trained language models, prompts in the form of natural language tokens or learnable word vectors have been designed to give different hints to different tasks, but it is less apparent what form prompts on graphs should take.  
Hence, \emph{how do we design prompts on graphs}, so that they can guide different downstream tasks to effectively make use of the pre-trained model?

\stitle{Present work: \model.}
To address these challenges, we propose a novel graph pre-training and prompting framework, called  \model, aiming to unify the pre-training and downstream tasks for GNNs.
Drawing inspiration from the prompting strategy for pre-trained language models, \model\ capitalizes on a unified template to define the objectives for both pre-training and downstream tasks, thus bridging their gap. We further equip \model\ with task-specific learnable prompts, which guides the downstream task to exploit relevant knowledge from the pre-trained GNN model. 
The unified approach endows \model\ with the ability of working on limited supervision such as few-shot learning tasks.

More specifically, to address the first challenge of unification, we focus on graph topology, which is a key enabler of graph models. In particular, subgraph is a universal structure that can be leveraged for both node- and graph-level tasks. At the node level, the information of a node can be enriched and represented by its contextual subgraph, \ie, a subgraph where the node resides in \cite{zhang2018link,huang2020graph}; at the graph level, the information of a graph is naturally represented by the maximum subgraph (\ie, the graph itself).
Consequently, we unify both the node- and graph-level tasks, whether in pre-training or downstream, into the same template: the similarity calculation of (sub)graph.  
In this work, we adopt link prediction as the self-supervised pre-training task, given that links are readily available in any graph without additional annotation cost. Meanwhile, we focus on the popular node classification and graph classification as downstream tasks, which are node- and graph-level tasks, respectively. All these tasks can be cast as instances of learning subgraph similarity. On one hand, the link prediction task in pre-training boils down to the similarity  between the contextual subgraphs of two nodes. On the other hand, the downstream node or graph classification task boils down to the similarity between the target instance (a node's contextual subgraph or the whole graph, resp.) and the class prototypical subgraphs constructed from labeled data. The unified template bridges the gap between the pre-training and different downstream tasks. 

Toward the second challenge, we distinguish different downstream tasks by way of the $\textsc{ReadOut}$ operation on subgraphs. The $\textsc{ReadOut}$ operation is essentially an aggregation function to fuse node representations in the subgraph into a single subgraph representation. For instance, sum pooling, which sums the representations of all nodes in the subgraph, is a practical and popular scheme for $\textsc{ReadOut}$. However, different downstream tasks can benefit from different aggregation schemes for their $\textsc{ReadOut}$. In particular, node classification tends to focus on features that can contribute to the representation of the target node, while  graph classification tends to focus on features associated with the graph class.
Motivated by such differences, we propose a novel task-specific learnable prompt to guide the $\textsc{ReadOut}$ operation of each downstream task with an appropriate aggregation scheme.
The learnable prompt serves as the parameters of the $\textsc{ReadOut}$ operation of downstream tasks, and thus enables different aggregation functions on the subgraphs of different tasks. 
Hence, \model\ not only unifies the pre-training and downstream tasks into the same template based on subgraph similarity, but also recognizes the differences between various downstream tasks to guide task-specific objectives.

\stitle{Extension to \nmodel.}
Although \model\ bridges the gap between pre-training and downstream tasks, we further propose a more generalized extension called  \nmodel\ to enhance the versatility of the unification framework. 
Recall that in \model, during the pre-training stage, a simple link prediction-based task is employed that naturally fits the subgraph similarity-based task template. Concurrently, in the prompt-tuning stage, the model integrates a learnable prompt vector only in the \textsc{ReadOut} layer of the pre-trained graph model. While these simple designs are effective, we take the opportunity to further raise two significant research questions.

First, toward a universal ``pre-train, prompt'' paradigm, it is essential to extend beyond a basic link prediction task in pre-training. 
While our proposed template can unify link prediction with typical downstream tasks on graph, researchers have proposed many other more advanced pre-training tasks on graphs, such as DGI \cite{velickovic2019deep} and GraphCL \cite{you2020graph}, which are able to capture more complex patterns from the pre-training graphs. Thus, to improve the compatibility of our framework with alternative pre-training tasks, a natural question arises: \emph{How can we unify a broader array of pre-training tasks within our framework?} 
In addressing this research question, we show that how a standard contrastive learning-based pre-training task on graphs can be generalized to fit our proposed task template, using a \emph{generalized pre-training loss}. The generalization anchors on the core idea of subgraph similarity within our task template, while preserves the established sampling strategies of the original pre-training task. Hence, the uniqueness of each pre-training task is still retained while ensuring compatibility with our task template. 
We further give generalized variants of popular graph pre-training tasks including DGI \cite{velickovic2019deep}, InfoGraph \cite{Sun2020InfoGraph}, GraphCL \cite{you2020graph}, GCC \cite{qiu2020gcc}.

Second, graph neural networks \cite{kipf2016semi,hamilton2017inductive,velivckovic2017graph,xu2018powerful} 
typically employ a hierarchical architecture in which each layer learns distinct knowledge in a certain aspect or at some resolution. For instance, the initial layers primarily processes raw node features, thereby focusing on the intrinsic properties of individual nodes. However, as the number of layers in the graph encoder increases, the receptive field has been progressively enlarged to process more extensive neighborhood data, thereby shifting the focus toward subgraph or graph-level knowledge. Not surprisingly, different downstream tasks may prioritize the knowledge encoded at different layers of the graph encoder. Consequently, it is important to generalize our prompt design to leverage the layer-wise hierarchical knowledge from the pre-trained encoder, beyond just the \textsc{ReadOut} layer in \model. Thus,  \emph{how do we design prompts that can adapt diverse downstream tasks to the hierarchical knowledge within multiple layers of the pre-trained graph encoder?}
To solve this question, we extend \model\ with a generalized prompt design. More specifically, instead of a single prompt vector applied to the \textsc{ReadOut} layer, we propose \emph{layer-wise prompts}, a series of learnable prompts that are integrated into each layer of the graph encoder in parallel. The series of prompts modifies all layers of the graph encoder (including the input and hidden layers), so that each prompt vector is able to locate the layer-specific knowledge most relevant to a downstream task, such as node-level characteristics, and local or global structural patterns.

\stitle{Contributions.}
To summarize, our contributions are fourfold.
(1) We recognize the gap between graph pre-training and downstream tasks, and propose a unification framework \model\ and its extension \nmodel based on subgraph similarity for both pre-training and downstream tasks, including both node and graph classification tasks.
(2) We propose a novel prompting strategy for \model, hinging on a learnable prompt to actively guide downstream tasks using task-specific aggregation in the $\textsc{ReadOut}$ layer, in order to drive the downstream tasks to exploit the pre-trained model in a task-specific manner.
(3) We extend \model\ to \nmodel, which further unifies existing popular graph pre-training tasks for compatibility with our task template, and generalizes the prompt design to capture the  hierarchical knowledge within each layer of the pre-trained graph encoder.
(4) We conduct extensive experiments on five public datasets, and the results demonstrate the superior performance of \model\ and \nmodel in comparison to the state-of-the-art approaches. 

A preliminary version of this manuscript has been published as a conference paper in The ACM Web Conference 2023 \cite{liu2023graphprompt}. We highlight the major changes as follows.
(1) Introduction: We reorganized Section~\ref{sec:intro} to highlight the motivation, challenges, and insights for the extension from \model\ to \nmodel. 
(2) Methodology: We proposed the extension \nmodel\ to allow more general pre-training tasks and prompt tuning in Section~\ref{sec:extend-model}. The proposed \nmodel\ not only broadens the scope of our task template to accommodate any standard contrastive pre-training task on graphs, but also enhances the extraction of hierarchical knowledge from the pre-trained graph encoder with layer-wise prompts.
(3) Experiments: We conducted additional experiments to evaluate the extended framework \nmodel\ in Section~\ref{sec:expt:perf:extend}, which demonstrate significant improvements over \model.

\section{Related Work}\label{related}

\stitle{Graph representation learning.}
The rise of graph representation learning, including earlier graph embedding \cite{perozzi2014deepwalk,tang2015line,grover2016node2vec} and recent GNNs \cite{kipf2016semi,hamilton2017inductive,velivckovic2017graph,xu2018powerful,jiang2023uncertainty,bo2022specformer,fang2024exgc,li2024graph}, opens up great opportunities for various downstream tasks at node and graph levels.
Note that learning graph-level representations requires an additional $\textsc{ReadOut}$ operation, which summarizes the global information of a graph by aggregating node representations through a flat \cite{duvenaud2015convolutional,gilmer2017neural,zhang2018end,xu2018powerful,jiang2024incomplete} or hierarchical \cite{gao2019graph,lee2019self,ma2019graph,ying2018hierarchical,bo2021beyond,li2024generalized,wu2024feasibility} pooling algorithm.

\stitle{Graph pre-training.}
Inspired by the application of pre-training models in language \cite{dong2019unified,beltagy2019scibert} and vision \cite{lu2019vilbert,bao2022beit} domains, graph pre-training \cite{xia2022survey} emerges as a powerful paradigm that leverages self-supervision on label-free graphs to learn intrinsic graph properties. 
While the pre-training learns a task-agnostic prior, a relatively light-weight fine-tuning step is further employed to update the pre-trained weights to fit a given downstream task. 
Different pre-training approaches design different self-supervised tasks based on various graph properties such as node features \cite{Hu2020Strategies,hu2020gpt}, links \cite{kipf2016variational,hamilton2017inductive,hu2020gpt,lu2021learning}, local or global patterns \cite{qiu2020gcc,Hu2020Strategies,lu2021learning}, local-global consistency \cite{velickovic2019deep,Sun2020InfoGraph},
and their combinations \cite{you2020graph,you2021graph,suresh2021adversarial}. 

However, the above approaches do not consider the gap between pre-training and downstream objectives, which limits their generalization ability to handle different tasks. Some recent studies recognize the importance of narrowing this gap. L2P-GNN \cite{lu2021learning} capitalizes on meta-learning \cite{finn2017model} to simulate the fine-tuning step during pre-training. However, since the downstream tasks can still differ from the simulation task, the problem is not fundamentally addressed.

\stitle{Prompt-based learning.}
In other fields, as an alternative to fine-tuning, researchers turn to prompting \cite{brown2020language}, in which a task-specific prompt is used to cue the downstream tasks. Prompts can be either handcrafted \cite{brown2020language} or learnable  \cite{liu2021gpt,lester2021power}. On graph data, the study of prompting is still limited \cite{yu2024few}. 
One recent work called GPPT \cite{sun2022gppt} capitalizes on a sophisticated design of learnable prompts on graphs, but it only works with node classification, lacking an unification effort to accommodate other downstream tasks like graph classification.
VNT \cite{tan2023virtual} utilizes a pre-trained graph transformer as graph encoder and introduces virtual nodes as soft prompts within the embedding space. However, same as GPPT, its application is only focused on the node classification task.
On the other hand, ProG \cite{sun2023all} and SGL-PT \cite{zhu2023sgl} both utilize a specific pre-training method for prompt-based learning on graphs across multiple downstream tasks. Additionally, HGPrompt \cite{yu2023hgprompt} and DyGPrompt \cite{yu2024dygprompt} extend GraphPrompt to address heterogeneous graph learning and dynamic graph learning respectively. Furthermore, MultiGPrompt \cite{yu2023multigprompt} proposes a multi-task pre-training and prompting framework. Moreover, MDGPT \cite{yu2024text} solve the multi-domain pre-training problem. Despite this, they neither explore the unification of different pre-training tasks, nor exploit the hierarchical knowledge inherent in graph encoder. Besides, there is a model also named as GraphPrompt \cite{zhang2021graphprompt}, but it considers an NLP task (biomedical entity normalization) on text data, where graph is only auxiliary. It employs the standard text prompt unified by masked language modeling, assisted by a relational graph to generate text templates, which is distinct from our work.

\stitle{Comparison to other settings.}
Our few-shot setting is different from other paradigms that also deal with label scarcity, including semi-supervised learning \cite{kipf2016semi} and meta-learning \cite{finn2017model}. In particular, semi-supervised learning cannot cope with novel classes not seen in training, while meta-learning requires a large volume of labeled data in their base classes for a meta-training phase, before they can handle few-shot tasks in testing. 

\section{Preliminaries} \label{sec:preliminaries}

In this section, we give the problem definition and introduce the background of GNNs.

\subsection{Problem Definition}\label{sec:prelim:problem}

\stitle{Graph.}
A graph can be defined as $G=(V,E)$, where $V$ is the set of nodes and $E$ is the set of edges. 
Equivalently, the graph can be represented by an adjacency matrix \vec{A}, such as $\vec{A}_{ij}=1$ iff $(v_i,v_j) \in E$, for any $v_i,v_j\in V$. 
We also assume an input feature matrix of the nodes, $\vec{X}\in\mathbb{R}^{|V|\times d}$, is available. Let $\vec{x}_i\in\mathbb{R}^d$ denote the feature vector of node $v_i\in V$. In addition, we denote a set of graphs as $\bG=\{G_1, G_2, \ldots, G_N\}$.

\stitle{Problem.}
In this paper, we investigate the problem of graph pre-training and prompting. For the downstream tasks, we consider the popular node classification and graph classification tasks.
For node classification on a graph $G=(V,E)$, let $C$ be the set of node classes with $\ell_i\in C$ denoting the class label of node $v_i \in V$.
For graph classification on a set of graphs $\bG$, let $\bC$ be the set of graph labels with $L_i\in \bC$ denoting the class label of graph $G_i\in \bG$.

In particular, the downstream tasks are given limited supervision in a few-shot setting: for each class in the two tasks, only $k$ labeled samples (\ie, nodes or graphs) are provided, known as $k$-shot classification.

\subsection{Graph Neural Networks}\label{sec:prelim:gnn}
The success of GNNs boils down to the message-passing mechanism \cite{wu2020comprehensive}, 
in which each node receives and aggregates messages (\ie, features or embeddings) from its neighboring nodes to generate its own representation. This operation of neighborhood aggregation can be stacked in multiple layers to enable recursive message passing.  In the $l$-th GNN layer, the embedding of node $v$, denoted by $\vec{h}^l_v$, is calculated based on the embeddings in the previous layer, as follows:
\begin{equation}
    \vec{h}^l_v =\textstyle \textsc{Aggr}(\vec{h}^{l-1}_v, \{\vec{h}^{l-1}_u : u\in\bN_v\}; \theta^l),
\end{equation}
where $\bN_v$ is the set of neighboring nodes of $v$, $\theta^l$ is the learnable GNN parameters in layer $l$. $\textsc{Aggr}(\cdot)$ is the neighborhood aggregation function and can take various forms \cite{hamilton2017inductive,velivckovic2017graph,xu2018powerful}
Note that in the first layer, the input node embedding $\vec{h}^0_v$ can be initialized as the node features in $\vec{X}$. 
The total learnable GNN parameters can be denoted as $\Theta=\{\theta^1, \theta^2, \ldots\}$. For brevity, we simply denote the output node representations of the last layer as $\vec{h}_v$.

For brevity of notation, GNNs can also be described in a alternative, matrix-based format. 
Consider the embedding matrix at the \(l\)-th layer, denoted as $\vec{H}^l$, in which 
each row, \(\vec{h}_i^l\), denotes the embedding vector of node \(v_i\). The embedding matrix at the \(l\)-th layer is calculated based on the embedding matrix from the previous, \ie, \((l-1)\)-th, layer:
\begin{equation}\label{eq.gnn}
\vec{H}^l =\textstyle \textsc{Aggr}(\vec{H}^{l-1},\Vec{A};\theta^l).
\end{equation}
The initial embedding matrix \(\vec{H}^0\) is set to be the same as the input feature matrix, \ie, \(\vec{H}^0 = \vec{X}\). After encoding through all the GNN layers, we simply denote the output embedding matrix as \(\vec{H}\). For easy of reference, we further abstract the multi-layer encoding process as follows.
\begin{align}
    \vec{H} =\textstyle \textsc{GraphEncoder}(\vec{X},\Vec{A};\Theta).
\end{align}

\section{Proposed Approach: \model}

In this section, we present \model, starting with a unification framework for common graph tasks. Then, we introduce the pre-training and downstream phases.

\begin{figure*}[t]
\centering
\includegraphics[width=1\linewidth]{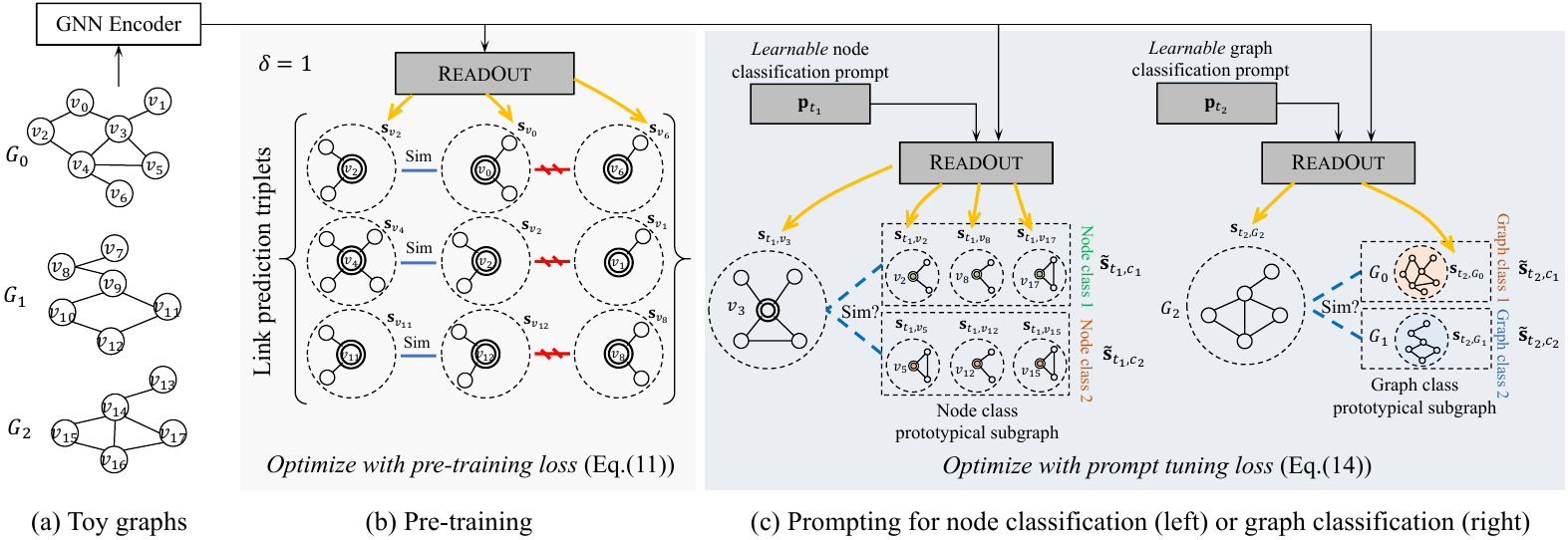}
\vspace{-6mm}
\caption{Overall framework of \model.}
\label{fig.framework}
\end{figure*}

\subsection{Unification Framework}\label{sec:model:unification}

We first introduce the overall framework of \model\ in Fig.~\ref{fig.framework}. Our framework is deployed on a set of label-free graphs shown in  Fig.~\ref{fig.framework}(a), for pre-training in  Fig.~\ref{fig.framework}(b). The pre-training adopts a link prediction task, which is self-supervised without requiring extra annotation. Afterward, in Fig.~\ref{fig.framework}(c), we capitalize on a learnable prompt to guide each downstream task, namely, node classification or graph classification, for task-specific exploitation of the pre-trained model.
We explain how the framework supports a unified view of pre-training and downstream tasks below.

\stitle{Instances as subgraphs.} The key to the unification of pre-training and downstream tasks lies in finding a common template for the tasks. The task-specific prompt can then be further fused with the template of each downstream task, to distinguish the varying characteristics of different tasks.

In comparison to other fields such as visual and language processing, 
graph learning is uniquely characterized by the exploitation of graph topology.
In particular, subgraph is a universal structure capable of expressing both node- and graph-level instances. 
On one hand, at the node level, every node resides in a local neighborhood, which in turn contextualizes the node \cite{DBLP:conf/cikm/LiuZFZH20,liu2021nodewise,liu2021tail}. The local neighborhood of a node $v$ on a graph $G=(V,E)$ is usually defined by a contextual subgraph $S_v=(V(S_v), E(S_v))$, where its set of nodes and edges are respectively given by
\begin{align}
V(S_v)&=\{d(u,v)\le \delta \mid u\in V\}, \text{ and }\\
E(S_v)&=\{ (u,u')\in E \mid u \in V(S_v), u'\in V(S_v) \},
\end{align}
where $d(u,v)$ gives the shortest distance between nodes $u$ and $v$ on the graph $G$, and $\delta$ is a predetermined threshold. That is, $S_v$ consists of nodes within $\delta$ hops from the node $v$, and the edges between those nodes. Thus, the contextual subgraph $S_v$ embodies not only the self-information of the node $v$, but also rich contextual information to complement the self-information \cite{zhang2018link,huang2020graph}.
On the other hand, at the graph level, the maximum subgraph of a graph $G$, denoted $S_{G}$, is the graph itself, \ie, $S_{G}=G$. The maximum subgraph  $S_{G}$ spontaneously embodies all information of $G$.
In summary, subgraphs can be used to represent both node- and graph-level instances: Given an instance $x$ which can either be a node or a graph (\eg, $x=v$ or $x=G$), the subgraph $S_x$ offers a unified access to the information associated with $x$.

\stitle{Unified task template.} Based on the above subgraph definitions for both node- and graph-level instances, we are ready to unify different tasks to follow a common template. Specifically, the link prediction task in pre-training and the downstream node and graph classification tasks can all be redefined as \emph{subgraph similarity learning}. Let $\vec{s}_x$ be the vector representation of the subgraph $S_x$, and $\text{sim}(\cdot,\cdot)$ be the cosine similarity function. As illustrated in Figs.~\ref{fig.framework}(b) and (c), the three tasks can be mapped to the computation  of subgraph similarity, which is formalized below. 
\begin{itemize}[leftmargin=*]
    \item \textbf{\textit{Link prediction}}: This is a node-level task. Given a graph $G=(V,E)$ and a triplet of nodes $(v,a,b)$ such that $(v,a)\in E$ and $(v,b)\notin E$, we shall have
    \begin{align}
    \text{sim}(\vec{s}_v,\vec{s}_a)>\text{sim}(\vec{s}_v,\vec{s}_b).
    \end{align}
    Intuitively, the contextual subgraph of $v$ shall be more similar to that of a node linked to $v$ than that of another unlinked node.
    
    \item \textbf{\textit{Node classification}}: This is also a node-level task. Consider a graph $G=(V,E)$ with a set of  node classes $C$, and a set of labeled nodes $D=\{(v_1,\ell_1),(v_2,\ell_2),\ldots\}$ where $v_i \in V$ and  $\ell_i$ is the corresponding label of $v_i$. As we adopt a $k$-shot setting, there are exactly $k$ pairs of $(v_i,\ell_i=c) \in D$ for every class $c\in C$.  
    For each class $c \in C$, further define a \emph{node class prototypical subgraph} represented by a vector $\vec{\tilde{s}}_c$, given by  
    \begin{align}\label{eq:nc-prototype}
        \vec{\tilde{s}}_c = \frac{1}{k}\sum_{(v_i,\ell_i)\in D, \ell_i=c} \vec{s}_{v_i}.
    \end{align}
    Note that the class prototypical subgraph is a ``virtual'' subgraph with a latent representation in the same embedding space as the contextual subgraphs. It is simply constructed as the mean representation of the contextual subgraphs of labeled nodes in a given class. 
    Then, given a node $v_j$ not in the labeled set $D$, its class label $\ell_j$ shall be 
    \begin{align}
        \ell_j =\textstyle \arg\max_{c\in C} \text{sim}(\vec{s}_{v_j}, \vec{\tilde{s}}_c).
    \end{align}
    Intuitively, a node shall belong to the class whose prototypical subgraph is the most similar to its contextual subgraph.
    
    \item \textbf{\textit{Graph classification}}: This is a graph-level task. Consider a set of graphs $\bG$ with a set of graph classes $\bC$, and a set of labeled graphs $\bD=\{(G_1,L_1),(G_2,L_2),\ldots\}$ where $G_i \in \bG$ and  $L_i$ is the corresponding label of $G_i$. In the $k$-shot setting, there are exactly $k$ pairs of $(G_i,L_i=c) \in \bD$ for every class $c\in \bC$.  
    Similar to node classification, for each class $c \in \bC$, we define a \emph{graph class prototypical subgraph}, also represented by the mean embedding vector of the (sub)graphs in $c$:
    \begin{align}\label{eq:gc-prototype}
        \vec{\tilde{s}}_c =\textstyle \frac{1}{k}\sum_{(G_i,L_i)\in \bD, L_i=c} \vec{s}_{G_i}.
    \end{align}
    Then, given a graph $G_j$ not in the labeled set $\bD$, its class label $L_j$ shall be 
    \begin{align}
        L_j =\textstyle \arg\max_{c\in \bC} \text{sim}(\vec{s}_{G_j}, \vec{\tilde{s}}_c).
    \end{align}
    Intuitively, a graph shall belong to the class whose prototypical subgraph is the most similar to itself. \hspace*{\fill}\IEEEQEDhere
\end{itemize}

It is worth noting that node and graph classification can be further condensed into a single set of notations. Let $(x,y)$ be an annotated instance of graph data, \ie, $x$ is either a node or a graph, and $y\in Y$ is the class label of $x$ among the set of classes $Y$. Then,  
\begin{align}
        y =\textstyle \arg\max_{c\in Y} \text{sim}(\vec{s}_{x}, \vec{\tilde{s}}_c).
    \end{align}

Finally, to materialize the common task template, we discuss how to learn the subgraph embedding vector $\vec{s}_x$ for the subgraph $S_x$. 
Given node representations $\vec{h}_{v}$ generated by a GNN (see Sect.~\ref{sec:prelim:gnn}), a standard approach of computing $\vec{s}_x$ is to employ a $\textsc{ReadOut}$ operation that aggregates the representations of nodes in the subgraph $S_x$. That is,
\begin{align} \label{eq.readout}
    \vec{s}_{x} =\textstyle \textsc{ReadOut}(\{\vec{h}_v:v\in V(S_x)\}).
\end{align}
The choice of the aggregation scheme for $\textsc{ReadOut}$ is flexible, including sum pooling and more advanced techniques \cite{ying2018hierarchical,xu2018powerful}. In our work, we simply use sum pooling.

In summary, the unification framework is enabled by the common task template of subgraph similarity learning, which lays the foundation of our pre-training and prompting strategies as we will introduce in the following parts.

\subsection{Pre-Training Phase}
As discussed earlier, our pre-training phase employs the link prediction task. Using link prediction/generation is a popular and natural way \cite{hamilton2017inductive,hu2020gpt,DBLP:conf/nips/HwangPKKHK20,lu2021learning},
as a vast number of links are readily available on large-scale graph data without extra annotation. In other words, the link prediction objective can be optimized on label-free graphs, such as those shown in Fig.~\ref{fig.framework}(a), in a self-supervised manner.  

Based on the common template defined in Sect.~\ref{sec:model:unification}, the link prediction task is anchored on the similarity of the contextual subgraphs of two candidate nodes. Generally, the subgraphs of two positive (\ie, linked) candidates shall be more similar than those of negative (\ie, non-linked) candidates, as illustrated in Fig.~\ref{fig.framework}(b).
Subsequently, the pre-trained prior on subgraph similarity can be naturally transferred to node classification downstream, which shares a similar intuition: The subgraphs of nodes in the same class shall be more similar than those of nodes from different classes. On the other hand, the prior can also support graph classification downstream, as graph similarity is consistent with subgraph similarity not only in letter (as a graph is technically always a subgraph of itself), but also in spirit. The ``spirit'' here refers to the tendency that graphs sharing similar subgraphs are likely to be similar themselves, which means graph similarity can be translated into the similarity of the containing subgraphs \cite{shervashidze2011weisfeiler,zhang2018end,togninalli2019wasserstein}.   

Formally, given a node $v$ on graph $G$, we randomly sample one positive node $a$ from $v$'s neighbors, and a negative node $b$ from the graph that does not link to $v$, forming a triplet $(v,a,b)$. Our objective is to increase the similarity between the contextual subgraphs $S_v$ and $S_a$, while decreasing that between $S_v$ and $S_b$. More generally, on a set of label-free graphs $\bG$, we sample a number of triplets from each graph to construct an overall training set $\bT_\text{pre}$. Then, we define the following pre-training loss.
\begin{align} \label{eq.pre-train-loss}
    \bL_{\text{pre}}(\Theta)=\textstyle -\sum_{(v,a,b)\in\bT_\text{pre}}\ln\frac{\exp(\text{sim}(\vec{s}_v,\vec{s}_a)/\tau)}{\sum_{u\in\{a,b\}}\exp(\text{sim}(\vec{s}_v,\vec{s}_u)/\tau)},
\end{align}
where $\tau$ is a temperature hyperparameter to control the shape of the output distribution. Note that the loss is parameterized by $\Theta$, which represents the GNN model weights. 

The output of the pre-training phase is the optimal model parameters $\Theta_0=\arg\min_\Theta \bL_{\text{pre}}(\Theta)$. $\Theta_0$ can be used to initialize the GNN weights for downstream tasks, thus enabling the transfer of prior knowledge downstream.

\subsection{Prompting for Downstream Tasks}
The unification of pre-training and downstream tasks enables more effective knowledge transfer as the tasks in the two phases are made more compatible by following a common template. However, it is still important to distinguish different downstream tasks, in order to capture task individuality and achieve task-specific optimum.

To cope with this challenge, we propose a novel task-specific learnable prompt on graphs, 
inspired by prompting in natural language processing \cite{brown2020language}. In language contexts, a prompt is initially a handcrafted instruction to guide the downstream task, which provides task-specific cues to extract relevant prior knowledge through a unified task template (typically, pre-training and downstream tasks are all mapped to masked language modeling). More recently, learnable prompts \cite{liu2021gpt,lester2021power} have been proposed as an alternative to handcrafted prompts, to alleviate the high engineering cost of the latter. 

\stitle{Prompt design.} Nevertheless, our proposal is distinctive from language-based prompting for two reasons.
Firstly, we have a different task template from masked language modeling. Secondly, since our prompts are designed for graph structures, they are more abstract and cannot take the form of language-based instructions. Thus, they are virtually impossible to be handcrafted. Instead, they should be topology related to align with the core of graph learning. In particular, under the same task template of subgraph similarity learning, 
the $\textsc{ReadOut}$ operation (used to generate the subgraph representation) can be ``prompted'' differently for different downstream tasks. Intuitively, different tasks can benefit
from different aggregation schemes for their $\textsc{ReadOut}$. For instance, node classification pays more attention to features that are topically more relevant to  the target node. In contrast, graph classification tends to focus on features that are correlated to the graph class. Moreover, the important features may also vary given different sets of instances or classes in a task.

Let $\vec{p}_t$ denote a learnable \emph{prompt vector} for a downstream task $t$, as shown in Fig.~\ref{fig.framework}(c). The prompt-assisted $\textsc{ReadOut}$ operation on a subgraph $S_x$ for task $t$ is
\begin{align}\label{eq:prompt-fw}
\vec{s}_{t,x} =\textstyle \textsc{ReadOut}(\{\vec{p}_t\odot\vec{h}_v:v\in V(S_x)\}),
\end{align}
where $\vec{s}_{t,x}$ is the task $t$-specific subgraph representation, and $\odot$ denotes the element-wise multiplication. That is, we perform a \emph{feature weighted} summation of the node representations from the subgraph, where the prompt vector $\vec{p}_t$ is a dimension-wise reweighting in order to extract the most relevant prior knowledge for the task $t$.

\stitle{Prompt tuning.}
To optimize the learnable prompt, also known as \emph{prompt tuning}, we formulate the loss based on the common template of subgraph similarity, using the prompt-assisted task-specific subgraph representations. 

Formally, consider a task $t$ with a labeled training set $\bT_{t} =\{(x_1,y_1),(x_2,y_2),\ldots\}$, where $x_i$ is an instance (\ie, a node or a graph), and $y_i\in Y$ is the class label of $x_i$ among the set of classes $Y$. The loss for prompt tuning is defined as%
{\small
\begin{align}\label{eq:prompt-loss}
   \bL_{\text{prompt}}(\vec{p}_t)=\textstyle -\sum_{(x_i,y_i)\in \bT_t}\ln\frac{\exp(\text{sim}(\vec{s}_{t,x_i},\tilde{\vec{s}}_{t,y_i})/\tau)}{\sum_{c\in Y}\exp(\text{sim}(\vec{s}_{t,x_i},\tilde{\vec{s}}_{t,c})/\tau)},
\end{align}}%
where the class prototypical subgraph for class $c$ is represented by $\tilde{\vec{s}}_{t,c}$, which is also generated by the prompt-assisted, task-specific $\textsc{ReadOut}$.

Note that, the prompt tuning loss is only parameterized by the learnable prompt vector $\vec{p}_t$, without the GNN weights. Instead, the pre-trained GNN weights   $\Theta_0$ are frozen for downstream tasks, as no fine-tuning is necessary. This significantly decreases the number of parameters to be updated downstream, thus not only improving the computational efficiency of task learning and inference, but also reducing the reliance on labeled data.

We outline the main steps for prompt tuning in Algorithm~\ref{alg.prompt}. In short, we iterate through each downstream task to learn their prompt vectors one by one. In lines 3--5, we obtain the embedding of each node using the pre-trained graph encoder, where the pre-trained weights $\Theta_0$ will be frozen throughout the prompt-tuning process.
In lines 8--16, we optimize the prompt vectors. More specifically, we perform prompt-assisted readout (lines 10--11), and update the embeddings of the prototypical subgraphs based on the few-shot labeled data given in the task (lines 12--13). Note that the steps of updating prototypical subgraphs are only required for classification tasks.

\begin{algorithm}[tbp]
\small
\caption{\textsc{Prompt Tuning for \model}}
\label{alg.prompt}
\begin{algorithmic}[1]
    \Require Graph encoder with pre-trained parameters $\Theta_0$, a set of $n$
    downstream tasks $\{t_1,\ldots,t_n\}$.
    \Ensure Optimized prompt vectors $\bP=\{\Vec{p}_1,\ldots,\Vec{p}_n\}$. 
    \For{$i\leftarrow$ $1$ to $n$}
        \State \slash* Encoding graphs with pre-trained weights *\slash
            \For{each graph $G=(\vec{X},\vec{A})$ in task $t_i$}
                \State $\vec{H}\leftarrow \textsc{GraphEncoder}(\Vec{X},\Vec{A};\Theta_0)$ 
                \State Let $\vec{h}_v\leftarrow \vec{H}[v]$, where $v$ is a node in $G$
            \EndFor
        \State $\Vec{p}_i\leftarrow$ initialization 
        \While{not converged} 
            \State \slash* Prompt-assisted readout by Eq.~\eqref{eq:prompt-fw} *\slash
            \For{each instance $x$ in task $t_i$}
                \State $\vec{s}_{t_i,x}\leftarrow \textsc{ReadOut}(\{\Vec{p}_i\odot \Vec{h}_v:v\in V(S_x)\})$ 
            \EndFor
            \State \slash* Update prototypical subgraphs *\slash
            \For{each class $c$ in task $t_i$} 
            \State $\vec{\tilde{s}}_{t_i,c}\leftarrow \textsc{Mean}(\vec{s}_{t_i,x}: x \text{ belongs to class } c)$
            \EndFor
            \State \slash* Optimizing the prompt vector *\slash
            \State Calculate $\bL_\text{prompt}$ by Eq.~\eqref{eq:prompt-loss}
            \State Update $\Vec{p}_i$ by backpropagating  $\bL_\text{prompt}$
        \EndWhile    
    \EndFor
    \State \Return $\{\Vec{p}_1,\ldots,\Vec{p}_n\}$.
\end{algorithmic}
\end{algorithm}


\section{Extension to \nmodel}\label{sec:extend-model}
Next, we generalize the pre-training tasks and prompts in \model, extending it to \nmodel.

\subsection{Generalizing Pre-Training Tasks}\label{sec:model-pre}
Although our proposed task template in \model\ easily accommodate the link prediction task, which is a simple, effective and popular pre-training task on graphs, it is not immediately apparent how the task template can fit other more advanced pre-training tasks on graphs. 
This presents a significant limitation in \model: Our ``pre-train, prompt'' framework is confined to only one specific pre-training task, thereby precluding other pre-training tasks that can potentially learn more comprehensive knowledge.
To unify a broader range of pre-training tasks in \nmodel, we first show that any standard contrastive graph pre-training task can be generalized to leverage subgraph similarity, the pivotal element in our task template. Meanwhile, each contrastive task can still preserve its unique sampling approach for the positive and negative pairs. 
Next, we illustrate the generalization process using several mainstream contrastive pre-training tasks. 

\begin{table*}[tbp]
\centering
\caption{Materializing the target, positive and negative instances for mainstream contrastive graph pre-training approaches for the generalized loss in Eq.~\eqref{eq:generalized_loss}. We also include link prediction (LP), which is used by \model.
} \label{table.pre-train}
\vspace{-2mm}
\small
\resizebox{0.8\linewidth}{!}{
\begin{tabular}{@{}c|c|c|c|c@{}}
\toprule
 & {Target instance $o$} & {Positive instance $a$} &{Negative instance $b$} & {Loss}\\
\midrule
     LP \cite{liu2023graphprompt}
     &a node $v$
     &a node linked to $v$  
     &a node not linked to $v$
     &Eq.~\eqref{eq.pre-train-loss}\\ \midrule
     DGI \cite{velickovic2019deep}
     & a graph $G$
     & a node in $G$
     & a node in $G'$, a corrupted graph of $G$
     &{Eq.~\eqref{eq.dgi}}\\ \midrule
     {InfoGraph \cite{Sun2020InfoGraph}} 
     & a graph $G$
     & a node in $G$
     & a node in $G'\ne G$ 
     & {Eq.~\eqref{eq.infograph}} \\\midrule
     \multirow{2}*{GraphCL \cite{you2020graph}} 
     & an augmented graph $G_i$ from
     & an augmented graph $G_j$ from
     & an augmented graph $G'_j$ from 
     & \multirow{2}*{Eq.~\eqref{eq.graphcl}}\\ 
     &  a graph $G$ by strategy $i$
     &  a graph $G$ by strategy $j$
     &  a graph $G'\ne G$ by strategy $j$
     & \\\midrule
     \multirow{2}*{GCC \cite{qiu2020gcc}} 
     & a random walk induced subgraph
     & a random walk induced subgraph
     & a random walk induced subgraph 
     &\multirow{2}*{Eq.~\eqref{eq:gcc}}\\
     &$G_v^r$ from a node $v$'s $r$-egonet
     &$\tilde{G}_v^r\neq G_v^r$ from $v$'s $r$-egonet
     &$G_v^{r'}$ from $v$'s $r'$-egonet, $r'\neq r$
     &\\
 \bottomrule
\end{tabular}}
\end{table*}

\stitle{Unification of graph contrastive tasks.}
We formulate a general loss term for any standard contrastive pre-training task on graphs. The generalized loss is compatible with our proposed task template, \ie, it is based on the core idea of subgraph similarity. The rationale for this generalization centers around two reasons.

First, the key idea of contrastive learning boils down to bringing positively related instances closer, while pushing away negatively related ones in their latent space \cite{jaiswal2020survey}. To encapsulate this objective in a generalized loss, the definition of instances needs to be unified for diverse contrastive tasks on graphs, so that the distance or proximity between the instances can also be standardized. Not surprisingly, \emph{subgraphs} can serve as a unified definition for various types of instances on graphs, including nodes \cite{velickovic2019deep, you2020graph}, subgraphs \cite{lu2021learning, xu2021self} or the whole graphs \cite{Sun2020InfoGraph, xu2021self}, which are common forms of instance in contrastive learning on graphs. By treating these instances as subgraphs, the contrastive objective can be accomplished by calculating a similarity score between the subgraphs, so as to maximize the similarity between a positive pair of subgraphs whilst minimizing that between a negative pair.

Second, the general loss term should flexibly preserve the unique characteristics of different contrastive tasks, thereby enabling the capture of diverse forms of knowledge through pre-training.
Specifically, various contrastive approaches diverge in the sampling or generation of positive and negative pairs of instances. Consequently, in our generalized loss, the set of positive or negative pairs can be materialized differently to accommodate the requirements of each contrastive objective.

Given the above considerations, we first define a standard contrastive graph pre-training task. Consider a set of pre-training data $\bT_\text{pre}$, which consists of the target instances (or elements that can derive the target instances). For each target instance $o \in \bT_\text{pre}$, 
a contrastive task samples or constructs a set of positive instances $Pos_o$, as well as a set of negative instances $Neg_o$, both w.r.t.~$o$. Hence, $\{(o, a) : a\in Pos_o\}$ is the set of positive pairs involving the target instance $o$, such that the objective is to maximize the similarity between each pair of $o$ and $a$. On the other hand, $\{(o, b) : b\in Neg_o\}$ is the set of negative pairs, such that the objective is to minimize the similarity between each pair of $o$ and $b$. 

Then, we propose the following generalized loss that can be used for any standard contrastive task. 
\begin{equation}\label{eq:generalized_loss}
     \bL(\Theta)=\textstyle -\sum_{o\in \bT_\text{pre}}\ln\frac{\sum_{a\in Pos_o}\exp(\text{sim}(\vec{s}_{a}, \vec{s}_{o})/\tau)}{\sum_{b\in Neg_o}\exp(\text{sim}(\vec{s}_{b}, \vec{s}_{o})/\tau)},
\end{equation}
where $\vec{s}_o$ denotes the subgraph embedding of the target instance, $\vec{s}_a, \vec{s}_b$ are the subgraph embedding of positive and negative instances, respectively, and $\tau$ is a temperature hyperparameter. 

Using this generalized loss, \nmodel\ broadens the scope of our proposed task template that is based on subgraph similarity, which can serve as a more general framework than \model\ to unify standard contrastive pre-training tasks on graphs beyond link prediction.

\stitle{Template for mainstream contrastive tasks.}
We further illustrate how the generalized loss term can be materialized for several mainstream contrastive graph pre-training approaches, in order to fit the task template. 
In Table~\ref{table.pre-train}, we summarize how various definitions of instances in these mainstream approaches are materialized, as also detailed below. Note that  the  strategies for sampling positive and negative instances are preserved as originally established.

\textbf{DGI} \cite{velickovic2019deep} operates on the principle of mutual information maximization, where the primary objective is to maximize the consistency between the local node representations and the global graph representation. 
Given a target instance $G\in \bT_\text{pre}$, a positive instance is any node $a$ in $G$, while a negative instance is any node $b$ in $G'$, a corrupted version of $G$. Then, we can reformulate the pre-training loss of DGI as 
\begin{align}\label{eq.dgi}
 \bL_{\text{DGI}}(\Theta)= \textstyle-\sum_{G\in \bT_\text{pre}}\ln\frac{\sum_{a\in V(G)}\exp\left(\text{sim}(\vec{s}_a, \vec{s}_G) / \tau \right)}{\sum_{b\in V(G')}\exp\left(\text{sim}(\vec{s}_b, \vec{s}_G) / \tau \right)},
\end{align}
where $V(G)$ denotes the set of nodes in $G$, and $G'$ is obtained by corrupting $G$.

\begin{figure}[t]
\centering
\includegraphics[width=1\linewidth]{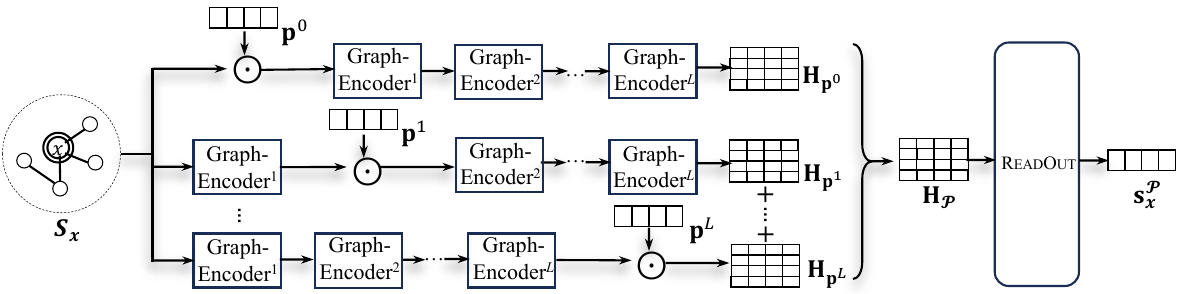}
\caption{Layer-wise prompt tuning for a pre-trained graph encoder with $L$ layers. \textsc{GraphEncoder}$^l$ represents the $l$-th layer of the graph encoder, and $\Vec{p}^{l}$ represents the prompt vector that modifies the $l$-th layer.}
\label{fig.prompt}
\end{figure}

\textbf{InfoGraph} \cite{Sun2020InfoGraph} extends the concept of DGI, which also attempts to maximize the mutual information between the global graph representation and local node representations, but differs from DGI in the sampling of negative instances.
Similar to DGI, given a target instance $G\in \bT_\text{pre}$, a positive instance is any node $a$ in $G$. However, a negative instance is a node sampled from a different graph $G'\neq G$ from the pre-training data, instead of a corrupted view of $G$ in DGI. 
Another difference from DGI is, the local representation of a node $v$ is derived by fusing the embeddings of $v$ from all layers of the graph encoder, an architecture that would not be affected by our reformulated loss as
{
\begin{align}\label{eq.infograph} \hspace{-1.5mm}
\bL_{\text{IG}}(\Theta)=-\sum_{G\in\bT_\text{pre}}\textstyle\ln\frac{\sum_{a\in V(G)}\exp(\text{sim}(\vec{s}_a,\vec{s}_G)/ \tau)}{\sum_{G'\in \bT_\text{pre}, G'\neq G}\sum_{b\in V(G')}\exp(\text{sim}(\vec{s}_b,\vec{s}_G)/ \tau)}.\hspace{-.5mm}
\end{align}}

\textbf{GraphCL} \cite{you2020graph} aims to maximize the mutual information between distinct augmentations of the same graph, while reducing that between augmentations of different graphs. Given a graph $G \in \bT_{pre}$, we can obtain distinct augmentations of $G$, say $G_i$ by augmentation strategy $i$ and $G_j$ by strategy $j$.
One of them, say $G_i$, can serve as a target instance, while the other, say $G_j$, can serve as a positive instance w.r.t.~$G_i$. Meanwhile, we can obtain an augmentation of a different graph $G'\ne G$, say $G'_j$, which can serve as a negative instance w.r.t.~$G_i$.
Thus, we can reformulate the pre-training loss as%
{
\begin{align}\label{eq.graphcl} \textstyle
    \bL_{\text{GCL}}(\Theta)= -\sum_{G\in \bT_\text{pre}}\ln \frac{\exp(\text{sim}(\vec{s}_{G_j} ,\vec{s}_{G_i} ) / \tau)}{\sum_{G'\in \bT_\text{pre}, G'\neq G}\exp(\text{sim}(\vec{s}_{G'_j},\vec{s}_{G_i}) / \tau)}.
\end{align}}%

\textbf{GCC} \cite{qiu2020gcc} employs a framework where the model learns to distinguish between subgraphs that are contextually similar or originated from the same root graph, and subgraphs that are derived from different root graphs. 
Given a node $v\in \bT_\text{pre}$, a target instance $G_v^r$ is a subgraph induced by performing random walk in the $r$-egonet of node $v$\footnote{The $r$-egonet of node $v$ is defined as a subgraph consisting of all nodes within $r$ hops of $v$ and all edges between these nodes}. To sample a positive instance, 
another random walk is performed in the same $r$-egonet, resulting in another induced subgraph $\tilde{G}_v^r$ that is different from $G_v^r$. 
On the other hand, a negative instance
can be sampled by random walk from the $r'$-egonet of $v$ such that $r'\ne r$. Letting $\bG_v^{r'}$ denote a set of random walk induced subgraphs from the $r'$-egonet of $v$ for some $r'\ne r$, we can reformulate the pre-training loss of GCC as 
\begin{align}\label{eq:gcc} \textstyle
\bL_{\text{GCC}}(\Theta)=-\sum_{v\in\bT_\text{pre}}\ln\frac{\exp(\text{sim}(\Vec{s}_{\tilde{G}_v^r}, \Vec{s}_{G_v^r} )/ \tau)}{\sum_{G_v^{r'}\in \bG_v^{r'}} \exp(\text{sim}(\Vec{s}_{G_v^{r'}}, \Vec{s}_{G_v^r})/ \tau )}.
\end{align}

\subsection{Generalizing Prompts}\label{sec:model-extension}

To adapt different downstream tasks more effectively, we propose a more generalized layer-wise prompt design to strategically utilize hierarchical knowledge across multiple layers of the pre-trained graph encoders.

\stitle{Layer-wise prompt design.}
Consider a pre-trained graph encoder comprising $L$ layers. 
In \nmodel, let $\bP$ be a set of $L+1$ prompt vectors, with one prompt vector allocated for each layer, including the input layer (\ie, $l=0$):
\begin{align} \textstyle
\bP=\{\mathbf{p}^{0}, \mathbf{p}^{1}, \ldots, \mathbf{p}^{L}\}.
\end{align}
That is, $\mathbf{p}^{l}$ is a learnable vector representing the prompt vector that modifies the $l$-th layer of the pre-trained encoder for the downstream task, for $0\le l\le L$. Note that in \model, there is only one prompt vector applied to the last layer $l=L$, which is then used by the \textsc{ReadOut} layer to obtain subgraph-level representations. In contrast, in \nmodel, these $L+1$ prompt vectors are applied to different layers in parallel in order to focus on different layers, as illustrated in Fig.~\ref{fig.prompt}.

Specifically, let $\vec{H}_{\mathbf{p}}$ denote the output from the pre-trained graph encoder after applying a single prompt vector $\vec{p}$ to a specific layer associated with $\vec{p}$, as follows.
\begin{align}\label{eq:token}
    \vec{H_\vec{p}} =\textstyle \textsc{GraphEncoder}_\vec{p}(\vec{X}, \vec{A}; \Theta),
\end{align}
where $\textsc{GraphEncoder}_\vec{p}(\cdot)$ indicates that a specific layer of the encoder has been modified by $\vec{p}$. Taking the prompt vector $\vec{p}^{l}$ as an example, it modifies the embedding matrix generated by the $l$-th layer of the graph encoder via element-wise multiplication, which gives $\vec{p}^{l}\odot \vec{H}^{l}$. Here $\vec{p}^{l}$ is multiplied element-wise with each row of $\vec{H}^l$ in a manner analogous to that in \model.  
Hence, the embedding matrix of the next layer is generated as follows:
\begin{align}\label{eq:modified_mp}
\vec{H}^{l+1} =\textstyle \textsc{Aggr}(\vec{p}^{l}\odot \vec{H}^{l},\Vec{A};\theta^{l+1}),
\end{align}
while the calculation of other layers remains unchanged, resulting in the output embedding matrix $\vec{H}_{\vec{p}^l}$.

Finally, we apply each prompt $\vec{p}^l \in \bP$ to the pre-trained graph encoder in parallel, and obtain a series of embedding matrices $\{\vec{H}_{\vec{p}^0},\vec{H}_{\vec{p}^1},\ldots,\vec{H}_{\vec{p}^L}\}$. 
We further fuse these ``post-prompt'' embedding matrices into an final output embedding matrix $\vec{H}_{\bP}$, which will be employed to calculate the downstream task loss. 
In particular, we adopt a learnable coefficient $w^l$ to weigh  $\vec{H}_{\vec{p}^l}$ in the fused output, \ie,
\begin{align}\label{eq:token-final}
\vec{H}_{\bP} =\textstyle \sum_{l=0}^L {w^l}\vec{H}_{\vec{p}^{l}},
\end{align}
as different downstream tasks may depend on information from diverse layers with varying degrees of importance.

Note that compared to \model, we suffer a $L$-fold increase in the number of learnable parameters during prompt tuning in \nmodel. However, a typical graph encoder adopts a shallow message-passing architecture with a small number of layers, \eg, $L=3$, which does not present a significant overhead.   

\stitle{Prompt tuning.}
Given a specific downstream task $t$, we have a set of task $t$-specific prompts $\bP_t=\{\vec{p}^{0}_t,\vec{p}^{1}_t,\ldots,\vec{p}^{L}_t\}$, corresponding to the $L+1$ layers in the graph encoder. After applying $\bP_t$ to the pre-trained encoder, we obtain the embedding for the subgraph $S_x$ as
\begin{align}\label{eq:tkde-prompt-fw}
\vec{s}_{t,x} =\textstyle \textsc{ReadOut}(\{\vec{h}_{\bP_t,v}:v\in V(S_x)\}),
\end{align}
where $\vec{s}_{t,x}$ is the task $t$-specific subgraph representation after applying the series of prompts $\bP_t$, and $\vec{h}_{\bP_t,v}$ is a row of $\vec{H}_{\bP_t}$ that corresponds to node $v$.

To optimize the generalized prompt vectors, we adopt the same loss function as utilized in \model, as given by Eq.~\eqref{eq:prompt-loss}. That is, we also leverage the task-specific subgraph representations after applying the generalized prompts, $\vec{s}_{t,x}$, in the prompt tuning loss.

\stitle{Complexity analysis.}
We compare the complexity of \model\ and \nmodel. On a downstream graph $G$, the computation consists of two main parts: Calculating the node embeddings with a pre-trained GNN, and prompt tuning. 
First, we calculate node embeddings with a pre-trained GNN, which is common to both \model\ and \nmodel, as well as any other method that utilizes a pre-trained GNN. This part depends on the complexity of the GNN. In a typical GNN, in each layer each node would access its $\bar{d}$ neighbors for aggregation, assuming at most $\bar{d}$ neighbors are involved in the aggregation. Hence, given $L$ layers, the overall complexity of calculating the node embeddings is $O(\bar{d}^L\cdot|V|)$, where $|V|$ is the number of nodes. 
Second, we perform prompt tuning. On one hand, for \model, a prompt vector modifies each node in $G$, leading to a complexity of $O(|V|)$. On the other hand, for \nmodel, $L+1$ prompt vectors modify each node in $G$, leading to a complexity of $O(L\cdot|V|)$.

Thus, for both \model\ and \nmodel, the first part is dominant in the complexity, since $O(\bar{d}^L\cdot|V|)$ is much larger than $O(|V|)$ and $O(L\cdot|V|)$. Hence, the prompt tuning step for both of our methods adds a negligible overhead.
Furthermore, comparing \nmodel\ to \model, the overhead is also manageable, since $L$ is often a small value, generally ranging from 1 to 3.


\section{Experiments}
In this section, we conduct extensive experiments including node classification and graph classification as downstream tasks on five benchmark datasets to evaluate \model.

\subsection{Experimental Setup}\label{sec:expt}

\stitle{Datasets.}
We employ five benchmark datasets for evaluation.
(1) \emph{Flickr} \cite{wen2021meta} is an image sharing network which is collected by SNAP\footnote{\url{https://snap.stanford.edu/data/}}. 
(2) \emph{PROTEINS} \cite{borgwardt2005protein} is a collection of protein graphs which include the amino acid sequence, conformation, structure, and features such as active sites of the proteins. 
(3) \emph{COX2} \cite{nr} is a dataset of molecular structures including 467 cyclooxygenase-2 inhibitors.
(4) \emph{ENZYMES} \cite{wang2022faith} is a dataset of 600 enzymes collected from BRENDA enzyme database. 
(5) \emph{BZR} \cite{nr} is a collection of 405 ligands for benzodiazepine receptor. 

We summarize these datasets in Table~\ref{table.datasets}.
Note that the ``Task'' column indicates the type of downstream task performed on each dataset: ``N'' for node classification and ``G'' for graph classification.

\begin{table}[tbp]
\center
\small
\caption{Summary of datasets. 
\label{table.datasets}}
\vspace{-1mm}
\resizebox{0.99\columnwidth}{!}{%
\begin{tabular}{@{}c|rrrrrrc@{}}
\toprule
	& \makecell[c]{ Graphs} &  \makecell[c]{Graph \\ classes} & \makecell[c]{Avg.\\ nodes} & \makecell[c]{Avg. \\ edges} &  \makecell[c]{Node \\ features} &  \makecell[c]{Node \\ classes} & \makecell[c]{Task \\ (N/G)} \\
\midrule
     Flickr & 1 & - & 89,250 & 899,756 & 500 & 7 & N \\ 
     PROTEINS & 1,113 & 2 & 39.06 & 72.82 & 1 & 3 & N, G\\
     COX2 & 467 & 2 & 41.22 & 43.45 & 3 & - & G\\
     ENZYMES & 600 & 6 & 32.63 & 62.14 & 18 & 3 & N, G\\
     BZR & 405 & 2 & 35.75 & 38.36 & 3 & - & G\\
 \bottomrule
\end{tabular}}
\vspace{-1mm}
\end{table}

\stitle{Baselines.}
We evaluate \model\ against the state-of-the-art approaches from three main categories.
(1) \emph{End-to-end graph neural networks}: GCN \cite{kipf2016semi}, GraphSAGE \cite{hamilton2017inductive}, GAT \cite{velivckovic2017graph} and GIN \cite{xu2018powerful}. They capitalize on the key operation of neighborhood aggregation to recursively aggregate messages from the neighbors, and work in an end-to-end manner.
(2) \emph{Graph pre-training models}: DGI \cite{velickovic2019deep}, InfoGraph \cite{Sun2020InfoGraph}, and GraphCL \cite{you2020graph}. They work in the ``pre-train, fine-tune'' paradigm. In particular, they pre-train the GNN models to preserve the intrinsic graph properties, and fine-tune the pre-trained weights on downstream tasks to fit task labels.
(3) \emph{Graph prompt models}: GPPT \cite{sun2022gppt}. GPPT utilizes a link prediction task for pre-training, and resorts to a learnable prompt for the node classification task, which is mapped to a link prediction task.

Other few-shot learning baselines on graphs, such as Meta-GNN \cite{zhou2019meta} and RALE \cite{liu2021relative}, often adopt a meta-learning paradigm \cite{finn2017model}. They cannot be used in our setting, as they require a large volume of labeled data in their base classes for the meta-training phase. In our approach, only label-free graphs are utilized for pre-training.

\stitle{Settings and parameters.}
To evaluate the goal of our \model\ in realizing a unified design that can suit different downstream tasks flexibly, we consider two typical types of downstream tasks, \ie, node classification and graph classification. 
In particular, for the datasets which are suitable for both of these two tasks, \ie, \emph{PROTEINS} and \emph{ENZYMES}, we only pre-train the GNN model once on each dataset, and utilize the same pre-trained model for the two downstream tasks with their task-specific prompting.

The downstream tasks follow a $k$-shot classification setting. For each type of downstream task, we construct a series of $k$-shot classification tasks. The details of task construction will be elaborated later when reporting the results in Sect.~\ref{sec:expt:perf}. For task evaluation, as the $k$-shot tasks are balanced classification, we employ accuracy as the evaluation metric following earlier work \cite{wang2020graph,liu2021relative}.

For all the baselines, based on the authors' code and default settings, we further tune their hyper-parameters to optimize their performance. More implementation details of the baselines and our approach can be found in Appendix D of GraphPrompt \cite{liu2023graphprompt}.

\subsection{Performance Evaluation}\label{sec:expt:perf}
As discussed, we perform two types of downstream task, namely, node classification and graph classification in few-shot settings. We first evaluate on a fixed-shot setting, and then vary the shot numbers to see the performance trend.

\stitle{Few-shot node classification.}
We conduct this node-level task on three datasets, \ie, \emph{Flickr}, \emph{PROTEINS}, and \emph{ENZYMES}. Following a typical $k$-shot setup \cite{zhou2019meta,wang2020graph,liu2021relative}, we generate a series of few-shot tasks for model training and validation.
In particular, for \emph{PROTEINS} and \emph{ENZYMES}, on each graph we randomly generate ten 1-shot node classification tasks (\ie, in each task, we randomly sample 1 node per class) for training and validation, respectively. Each training task is paired with a validation task, and the remaining nodes not sampled by the pair of training and validation tasks will be used for testing. For \emph{Flickr}, as it contains a large number of very sparse node features, selecting very few shots for training may result in inferior performance for all the methods.
Therefore, we randomly generate ten 50-shot node classification tasks, for training and validation, respectively. On Flickr, 50 shots are still considered few, accounting for less than 0.06\% of all nodes on the graph.  

Table~\ref{table.node-classification} illustrates the results of few-shot node classification. We have the following observations. 
First, our proposed \model\ outperforms all the baselines across the three datasets, demonstrating the effectiveness of \model\ in transferring knowledge from the pre-training to downstream tasks for superior performance.
In particular, by virtue of the unification framework and prompt-based task-specific aggregation in $\textsc{ReadOut}$ function, \model\ is able to close the gap between pre-training and downstream tasks, and derive the downstream tasks to exploit the pre-trained model in task-specific manner.
Second, compared to graph pre-training models, GNN models usually achieve comparable or even slightly better performance. This implies that the discrepancy between the pre-training and downstream tasks in these pre-training models, obstructs the knowledge transfer from the former to the latter. Even with sophisticated pre-training, they cannot effectively promote the performance of downstream tasks.
Third, the graph prompt model GPPT is only comparable to or even worse than the other baselines, despite also using prompts. A potential reason is that GPPT requires much more learnable parameters in their prompts than ours, which may not work well with very few shots. 

\begin{table}[tbp]
    \centering
    \small
     \addtolength{\tabcolsep}{-1mm}
    \caption{Accuracy evaluation on node classification.
    }
    \label{table.node-classification}%
    \vspace{-1mm} 
    {\footnotesize Results are in percent, with best \textbf{bolded} and runner-up \underline{underlined}.}
    \\[2mm] 
    \resizebox{0.8\linewidth}{!}{%
    \begin{tabular}{@{}l|c|c|c@{}}
    \toprule
  \multirow{2}*{Methods} & \multicolumn{1}{c|}{Flickr} & \multicolumn{1}{c|}{PROTEINS} & \multicolumn{1}{c}{ENZYMES}  \\ & 50-shot & 1-shot & 1-shot  \\\midrule\midrule
    \method{GCN} &  9.22 $\pm$ 9.49 & 59.60 $\pm$ 12.44 & 61.49 $\pm$ 12.87 \\
    \method{GraphSAGE}  & 13.52 $\pm$ 11.28 & 59.12 $\pm$ 12.14 & 61.81 $\pm$ 13.19\\
     \method{GAT} & 16.02 $\pm$ 12.72 & 58.14 $\pm$ 12.05 & 60.77 $\pm$ 13.21\\
    \method{GIN} & 10.18 $\pm$ 5.41 & \underline{60.53} $\pm$ 12.19  & \underline{63.81} $\pm$ 11.28 
\\\midrule
    \method{DGI} & 17.71 $\pm$ 1.09 & 54.92 $\pm$ 18.46 & 63.33 $\pm$ 18.13\\
    \method{GraphCL}  & 18.37 $\pm$ 1.72 & 52.00 $\pm$ 15.83 & 58.73 $\pm$ 16.47\\
    \midrule
    \method{GPPT} & \underline{18.95} $\pm$ 1.92 & 50.83 $\pm$ 16.56 & 53.79 $\pm$ 17.46\\
    \midrule
    \method{GraphPrompt} & \textbf{20.21} $\pm$ 11.52 & \textbf{63.03} $\pm$ 12.14 & \textbf{67.04} $\pm$ 11.48\\
    \bottomrule
        \end{tabular}}
    \vspace{-1mm}
\end{table}

\stitle{Few-shot graph classification.}
We further conduct few-shot graph classification on four datasets, \ie, \emph{PROTEINS}, \emph{COX2}, \emph{ENZYMES}, and \emph{BZR}. For each dataset, we randomly generate 100 5-shot classification tasks for training and validation, following a similar process for node classification tasks. 

We illustrate the results of few-shot graph classification in Table~\ref{table.graph-classification}, and have the following observations.
First, our proposed \model\ significantly outperforms the baselines on these four datasets. This again demonstrates the necessity of unification for pre-training and downstream tasks, and the effectiveness of prompt-assisted task-specific aggregation for $\textsc{ReadOut}$.
Second, as both node and graph classification tasks share the same pre-trained model on \emph{PROTEINS} and \emph{ENZYMES}, the superior performance of \model\ on both types of tasks further demonstrates that, the gap between different tasks is well addressed by virtue of our unification framework.
Third, the graph pre-training models generally achieve better performance than the end-to-end GNN models.
This is because both InfoGraph and GraphCL capitalize on graph-level tasks for pre-training, which are naturally not far away from the downstream graph classification.

\begin{table}[tbp] 
    \centering
    \small
     \addtolength{\tabcolsep}{-1mm}
    \caption{Accuracy evaluation on graph classification.}
    \vspace{-1mm}
    \label{table.graph-classification}%
    \resizebox{1.0\linewidth}{!}{%
    \begin{tabular}{@{}l|c|c|c|c@{}}
    \toprule
  \multirow{2}*{Methods} & \multicolumn{1}{c|}{PROTEINS} & \multicolumn{1}{c|}{COX2} & \multicolumn{1}{c|}{ENZYMES} & \multicolumn{1}{c}{BZR}  \\ & 5-shot  & 5-shot & 5-shot & 5-shot \\\midrule\midrule
    \method{GCN} & 54.87 $\pm$ 11.20 & 51.37 $\pm$ 11.06  & 20.37 $\pm$ 5.24	 & 56.16 $\pm$ 11.07	\\
    \method{GraphSAGE} & 52.99 $\pm$ 10.57 & 52.87 $\pm$ 11.46	 & 18.31 $\pm$ 6.22 & 57.23 $\pm$ 10.95\\
     \method{GAT} & 48.78 $\pm$ 18.46 & 51.20 $\pm$ 27.93 & 15.90 $\pm$ 4.13 & 53.19 $\pm$ 20.61\\
    \method{GIN} & \underline{58.17} $\pm$ 8.58 & 51.89 $\pm$ 8.71	& 20.34 $\pm$ 5.01 & 57.45 $\pm$ 10.54\\
    \midrule
    \method{InfoGraph} & 54.12 $\pm$ 8.20 & 54.04 $\pm$ 9.45 & 20.90 $\pm$ 3.32 & 57.57 $\pm$ 9.93\\
    \method{GraphCL} & 56.38 $\pm$ 7.24  & \underline{55.40} $\pm$ 12.04  & \underline{28.11} $\pm$ 4.00  & \underline{59.22} $\pm$ 7.42	\\
    \midrule
    \method{GraphPrompt} & \textbf{64.42} $\pm$ 4.37  & \textbf{59.21} $\pm$ 6.82 & \textbf{31.45} $\pm$ 4.32	 & \textbf{61.63} $\pm$ 7.68 \\\bottomrule
    \end{tabular}}
    \vspace{-1mm}
\end{table}

\begin{figure}[t]
\centering
\includegraphics[width=1.0\linewidth]{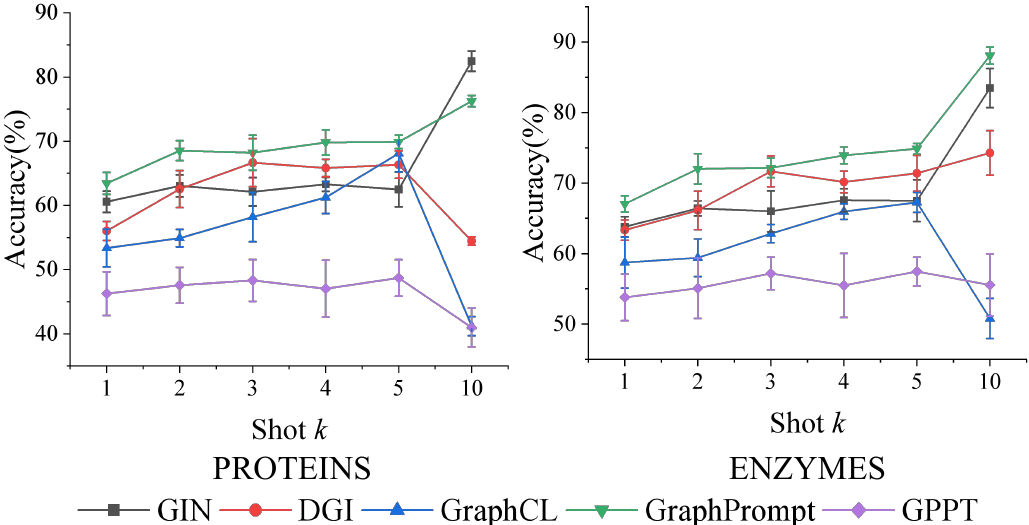}
\vspace{-2mm}
\caption{Impact of shots on few-shot node classification.}
\label{fig.node-few-shot-tune}
\end{figure}

\begin{figure}[t]
\centering
\includegraphics[width=1.0\linewidth]{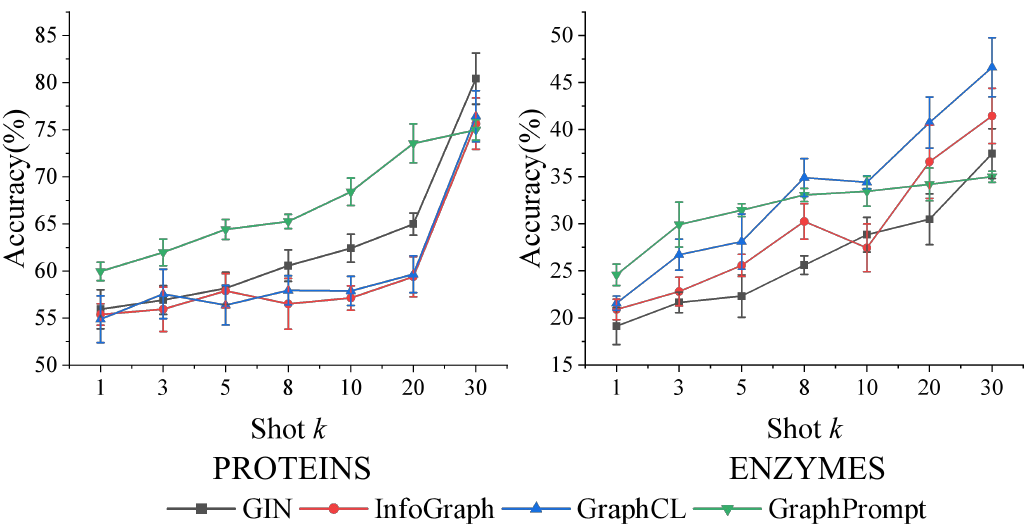}
\vspace{-2mm}
\caption{Impact of shots on few-shot graph classification.}
\label{fig.graph-few-shot-tune}
\end{figure}

\begin{figure}[t]
\centering
\includegraphics[width=1.0\linewidth]{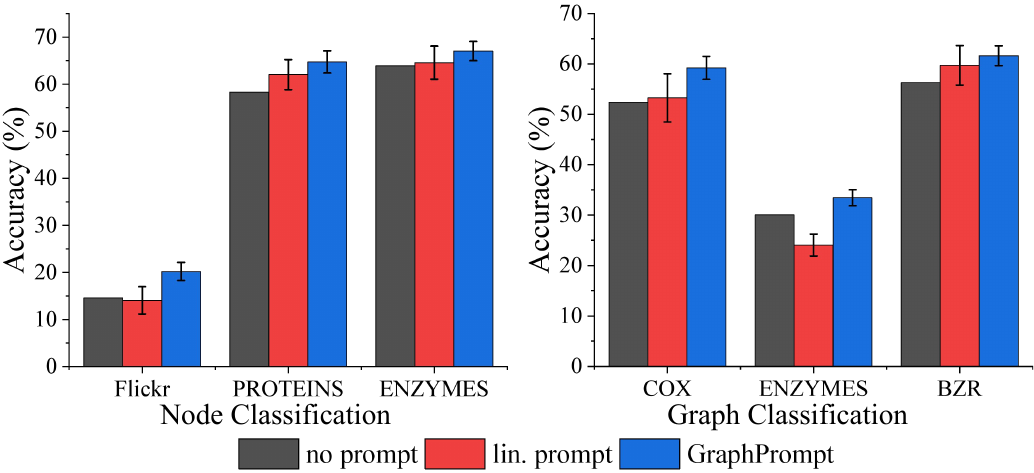}
\vspace{-2mm}
\caption{Ablation study.}
\label{fig.ablation}
\end{figure}

\stitle{Performance with different shots.}
We further tune the number of shots for the two few-shot classification tasks to evaluate its influence on the performance. In particular, for few-shot node classification, we tune the number of shots in \{1, 2, 3, 4, 5, 10\}, and employ the most competitive baselines (\ie, GIN, DGI, GraphCL, and GPPT) for comparison.
Similarly, for few-shot graph classification, we tune the number of shots in \{1, 3, 5, 8, 10, 20\}, and also employ the most competitive baselines (\eg, GIN, InfoGraph, and GraphCL) for comparison.
The number of tasks is identical to the above settings of few-shot node classification and graph classification. We conduct experiments on two datasets, \ie, \emph{PROTEINS} and \emph{ENZYMES} for the two tasks.

We illustrate the comparison in Figs.~\ref{fig.node-few-shot-tune} and \ref{fig.graph-few-shot-tune} for node and graph classification\footnote{\label{note:errbar}The error bars in the figures represent standard deviation.}, respectively, and have the following observations.
First, for few-shot node classification, in general our proposed \model\ consistently outperforms the baselines across different shots. The only exception occurs on \emph{PROTEINS} with 10-shot, which is possibly because 10-shot labeled data might be sufficient for GIN to work in an end-to-end manner.
Second, for few-shot graph classification, our proposed \model\ outperforms the baselines when very limited labeled data is given (\eg, when number of shots is 1, 3, or 5), and might be surpassed by some competitive baselines when fed with adequate labeled data (\eg, GraphCL with number of shot 8, 10, and 20).

\subsection{Ablation Study}
To evaluate the contribution of each component, we conduct an ablation study by comparing \model\ with different prompting strategies:
(1) \emph{no prompt}: for downstream tasks, we remove the prompt vector, and conduct classification by employing a classifier on the subgraph representations obtained by a direct sum-based $\textsc{ReadOut}$.
(2) \emph{lin. prompt}: we replace the prompt vector with a linear transformation matrix as an alternative design.

We report the comparison in Fig.~\ref{fig.ablation}, and note the following$^{\text{\ref{note:errbar}}}$.
(1) Without the prompt vector, \emph{no prompt} usually performs the worst in the four models, showing the necessity of prompting the $\textsc{ReadOut}$ operation differently for different downstream tasks.
(2) Converting the prompt vector into a linear transformation matrix also hurts the performance, as the matrix involves more parameters thus increasing the reliance on labeled data.

\section{Experiments on \nmodel}\label{sec:expt:perf:extend}

In this section, we further conduct experiments to evaluate our extended approach \nmodel\ in comparison to the vanilla \model.

We follow the same experiment setup of \model, as detailed in Sect.~\ref{sec:expt}. 
In particular, we evaluate the same few-shot node classification or graph classification tasks on the same five datasets.
Consistent with \model, in the implementation of \nmodel, we utilize a three-layer GIN as the backbone, and set the hidden dimensions as 32.  We also set $\delta=1$ to construct 1-hop subgraphs for the nodes same as \model. 

\subsection{Effect of Generalized Layer-wise Prompts}\label{sec:expt+:layer-prompt}
We first investigate the impact of layer-wise prompts to evaluate their ability of extracting hierarchical pre-trained knowledge.
To isolate the effect of prompts and be comparable to \model, we fix the pre-training task in \nmodel\ to link prediction; further experiments on alternative pre-training tasks will be presented in Sect.~\ref{sec:expt+:pre-training}.
Furthermore, we compare to three variants of \nmodel, namely, \nmodel/0, \nmodel/1 and \nmodel/2. Here,
\nmodel/$l$ denotes the variant where only one prompt vector $\vec{p}^l$ is applied to modify the $l$-th layer of the pre-trained encoder. As we have a total of $L=3$ layers, \model\ is equivalent to \nmodel/3.

In the following, we perform few-shot node and graph classification as the downstream tasks, and analyze the results of \model, \nmodel\ and the variants.

\stitle{Few-shot node classification.}
The results of node classification are reported in Table~\ref{table.node-classification-extend}. We first observe that \nmodel\ outperforms all variants and \model. Except \nmodel, only a single prompt vector is added to a specific layer of the graph encoder. This indicates the benefit of leveraging multiple prompt vectors in a layer-wise manner to extract hierarchical knowledge within the pre-trained graph encoder.
Second, the application of prompts at different layers leads to varying performance. This observation confirms the nuanced distinctions across the layers of the graph encoder, suggesting the existence of hierarchical structures within pre-trained knowledge.  
Third, the significance of prompt vectors at different layers varies across datasets. 
Specifically, for the \textit{Flickr} dataset, applying the prompt to a deeper layer generally yields progressively better performance. In contrast, for the \textit{PROTEINS}  and \textit{ENZYMES} datasets, implementing the prompt at the last layer (\ie, \model) is generally not as good compared to its application at the shallow layers. This discrepancy is attributed to the notable difference in the nature and size of the graphs. Note that the \textit{Flickr} graph is a relatively large image sharing network, while the graphs in the \textit{PROTEINS} and \textit{ENZYMES} datasets describe small chemical structures, as detailed in Table~\ref{table.datasets}. 
In smaller graphs, knowledge from shallower layers could be adequate, given the relative size of the receptive field to the graph. 
Nevertheless, \nmodel\ automatically selects the most relevant layers and achieves the best performance.

\stitle{Few-shot graph classification.}
The results for graph classification 
are presented in Table~\ref{table.graph-classification-extend}.
We first observe that \nmodel\ continues to outperform all variants and \model\ on graph classification, similar to the results on node classification. This demonstrates the robustness of our  layer-wise prompts on different types of graph tasks.

Moreover, compared to node classification, \nmodel\ shows a more pronounced improvement over the variants. 
%
This difference stems from the nature of graph classification as a graph-level task, which emphasizes the need for more global knowledge from all layers of the graph encoder. 
Thus, effective integration of hierarchical knowledge across these layers is more important to graph classification, which can enhance performance more significantly.

\begin{table}[tbp] 
    \centering
    \small
     \addtolength{\tabcolsep}{1mm}
    \caption{Effect of layer-wise prompts (node classification).
    }
    \label{table.node-classification-extend}%
    \vspace{-2mm} 
    \resizebox{1\linewidth}{!}{%
    \begin{tabular}{@{}l|c|c|c@{}}
    \toprule
  \multirow{2}*{Methods} & Flickr & PROTEINS & ENZYMES\\ 
  & 50-shot & 1-shot & 1-shot  \\\midrule\midrule
    \nmodel/0  & 17.18 $\pm$ 11.49 & 62.64 $\pm$ 13.31 & 68.32 $\pm$ 10.54\\
     \nmodel/1 & 18.14 $\pm$ 11.82 &   63.02    $\pm$    12.04  & 68.52 $\pm$ 10.77\\
    \nmodel/2 & 20.11 $\pm$ 12.58 & \underline{63.61} $\pm$ 11.89  & \underline{69.09} $\pm$ 10.19\\
    \model &  \underline{20.21} $\pm$ 11.52 & 63.03 $\pm$ 12.14 & 67.04 $\pm$ 11.48
\\\midrule
    \nmodel & \textbf{20.55} $\pm$ 11.97 &  \textbf{63.64} $\pm$ 11.69 & \textbf{69.28} $\pm$ 10.74\\
    ($\uparrow$~vs.~\model) &(+1.68\%) &(+0.97\%) &(+3.34\%) \\
    \bottomrule
        \end{tabular}}
\end{table}

\begin{table}[tbp] 
    \centering
    \small
     \addtolength{\tabcolsep}{1mm}
    \caption{Effect of layer-wise prompts (graph classification).}
    \vspace{-2mm}
    \label{table.graph-classification-extend}%
    \resizebox{1.0\linewidth}{!}{%
    \begin{tabular}{@{}l|c|c|c|c@{}}
    \toprule
  \multirow{2}*{Methods} & PROTEINS & COX2 & ENZYMES & BZR  \\ & 5-shot  & 5-shot & 5-shot & 5-shot \\\midrule\midrule
    \nmodel/0  & 60.80 $\pm$ 2.96 & 51.14 $\pm$ 12.93 & 44.93 $\pm$ 3.91& 53.93 $\pm$ 5.37\\
     \nmodel/1 & 56.49 $\pm$ 8.79 & 53.53 $\pm$ 6.73 & \underline{35.91} $\pm$ 4.64 & 48.50 $\pm$ 1.80\\
    \nmodel/2 & 61.63 $\pm$ 6.86 & 58.16 $\pm$ 6.97  & 34.36 $\pm$ 5.16 & 48.50 $\pm$ 2.43\\
    \model &  \underline{64.42} $\pm$ 4.37 & \underline{59.21} $\pm$ 6.82 & 31.45 $\pm$ 4.32  & \underline{61.63} $\pm$ 7.68
\\\midrule
    \nmodel & \textbf{67.71} $\pm$ 7.09 & \textbf{65.23} $\pm$ 5.59 & \textbf{45.35} $\pm$ 4.15& \textbf{68.61} $\pm$ 3.99\\
    ($\uparrow$~vs.~\model) &(+5.11\%) &(+10.17\%) &(+44.20\%) &(+11.33\%) \\
        \bottomrule
    \end{tabular}}
\end{table}

\subsection{Compatibility with Generalized Pre-training Tasks}\label{sec:expt+:pre-training}
Finally, we conduct experiments using alternative contrastive learning approaches for pre-training, beyond the simple link prediction task. Specifically, we select the two most popular contrastive pre-training task on graphs, \ie, DGI \cite{velickovic2019deep} and GraphCL \cite{you2020graph}, and implement the generalized pre-training loss for each of them as discussed in Sect.~\ref{sec:model-pre}. For each pre-training task, we compare among \nmodel, \model, and the original architecture in their paper without prompt tuning (denoted as ``Original''). 

The results of node and graph classification tasks are reported in Tables~\ref{table.node-classification-pre} and \ref{table.graph-classification-pre}, respectively. It is evident that both \nmodel\ and \model\ exhibit superior performance compared to the original versions of DGI and GraphCL\footnote{One exception occurs when using GraphCL as the pre-training method for the 5-shot graph classification task on \textit{PROTEINS}. The reason could be that 5-shot labeled data already provide adequate supervision in this case, but our methods are still  superior in more label-scarce scenarios. Specifically, in the one-shot scenario, the accuracy of original GraphCL, \model\ and \nmodel\ is 53.26, 53.77 and 54.07, respectively.}.
The results imply that our proposed prompt-based framework can flexibly incorporate well-known contrastive pre-training models like DGI and GraphCL, overcoming the limitation of a singular pre-training approach based on link prediction.
Furthermore, empirical results also reveal a consistent trend where \nmodel\ demonstrates superior performance over \model. This trend further corroborates the effectiveness of layer-wise prompts under alternative pre-training tasks, extending the analysis in Sect.~\ref{sec:expt+:layer-prompt}. 

\begin{table}[tbp] 
    \centering
    \small
     \addtolength{\tabcolsep}{1mm}
    \caption{Compatibility with popular contrastive pre-training on graphs, for downstream node classification.
    }
    \label{table.node-classification-pre}%
    \vspace{-2mm} 
    \resizebox{1\linewidth}{!}{%
    \begin{tabular}{@{}l|l|c|c|c@{}}
    \toprule
  \multirow{2}*{Pre-training} & \multirow{2}*{Methods} & Flickr & PROTEINS & ENZYMES  \\& & 50-shot & 1-shot & 1-shot  \\\midrule\midrule
    \multirow{3}*{DGI} 
    & Original & 17.71 $\pm$ 1.09 & 54.92 $\pm$ 18.46& 63.33 $\pm$ 18.13\\
    & \model & \underline{17.78} $\pm$ 4.98 & \underline{60.79} $\pm$ 12.00 & \underline{66.46} $\pm$ 11.39 \\
    & \nmodel & \textbf{20.98} $\pm$ 11.60 & \textbf{65.24} $\pm$ 12.51 & \textbf{68.92} $\pm$ 10.77 \\
    
\midrule
    \multirow{3}*{GraphCL} 
    & Original   & 18.37 $\pm$ 1.72 & 52.00 $\pm$ 15.83 & 58.73 $\pm$ 16.47 \\
    & \model & \underline{19.33} $\pm$ 4.11 & \underline{60.15} $\pm$ 13.31 & \underline{63.14} $\pm$ 11.02 \\
    & \nmodel & \textbf{19.95} $\pm$ 12.48 & \textbf{62.55} $\pm$ 12.63 & \textbf{68.17} $\pm$ 11.30 \\
    
    \bottomrule
        \end{tabular}}
\end{table}

\begin{table}[tbp] 
    \centering
    \small
     \addtolength{\tabcolsep}{1mm}
    \caption{Compatibility with popular contrastive pre-training on graphs, for downstream graph classification.}
    \vspace{-2mm}
    \label{table.graph-classification-pre}%
    \resizebox{1.0\linewidth}{!}{%
    \begin{tabular}{@{}l|l|c|c|c|c@{}}
    \toprule
  \multirow{2}*{Pre-training} & \multirow{2}*{Methods} & PROTEINS & COX2 & ENZYMES & BZR  \\& & 5-shot  & 5-shot & 5-shot & 5-shot \\\midrule\midrule
    \multirow{3}*{DGI} 
    & Original & 54.12 $\pm$ 8.20 & 54.04 $\pm$ 9.45 & 20.90$\pm$ 3.32 & 57.57$\pm$ 9.93\\
    & \model & \underline{54.32} $\pm$ 0.61 & \textbf{54.60} $\pm$ 5.96 &  \underline{27.69} $\pm$ 2.23 &  \underline{58.53} $\pm$ 7.41 \\
    & \nmodel & \textbf{56.16} $\pm$ 0.67 & \underline{54.30} $\pm$ 6.58 & \textbf{41.40} $\pm$ 3.77& \textbf{62.83} $\pm$ 8.97\\
\midrule
    \multirow{3}*{GraphCL} 
    & Original & \textbf{56.38} $\pm$ 7.24 & 55.40 $\pm$ 12.04 & 28.11 $\pm$ 4.00 & 59.22$\pm$ 7.42\\\
    & \model & 55.30 $\pm$ 1.05 & \underline{57.87} $\pm$ 1.52 &\underline{29.82} $\pm$ 2.87& \underline{61.35} $\pm$ 7.54\\
    & \nmodel & \underline{55.59} $\pm$ 2.63 & \textbf{59.32} $\pm$ 17.00 & \textbf{41.42} $\pm$ 3.73& \textbf{61.75} $\pm$ 7.56\\
        \bottomrule

    \end{tabular}}
\end{table}

\section{Conclusions}
In this paper, we studied the research problem of prompting on graphs and proposed \model, in order to overcome the limitations of graph neural networks in the supervised or ``pre-train, fine-tune'' paradigms.
In particular, to narrow the gap between pre-training and downstream objectives on graphs, we introduced a unification framework by mapping different tasks to a common task template. Moreover, to distinguish task individuality and achieve  task-specific optima, we proposed a learnable task-specific prompt vector that guides each downstream task to make full of the pre-trained model.  
We further extended \model\ into \nmodel by enhancing both the pre-training and prompt tuning stages.
Finally, we conduct extensive experiments on five public datasets, and show that \model\ 
and \nmodel
significantly outperforms various state-of-the-art baselines.

\section*{Acknowledgments}
This research / project is supported by the Ministry of Education, Singapore, under its Academic Research Fund Tier 2 (Proposal ID: T2EP20122-0041). Any opinions, findings and conclusions or recommendations expressed in this material are those of the author(s) and do not reflect the views of the Ministry of Education, Singapore. This work is also supported in part by the National Key Research and Development Program of China under Grant 2020YFB2103803. The first author extends his heartfelt gratitude to Ms.~Mengzhuo Fang for her invaluable support and assistance during challenging periods.


\bibliographystyle{IEEEtran}
\bibliography{references}

\begin{IEEEbiographynophoto}{Xingtong Yu}
received his PhD degree in computer science from the University of Science and Technology of China in 2024, and the bachelor’s degree from School of the Gifted Young, University of Science and Technology of China in 2019.  His current research focuses on graph-based machine learning, graph mining, and prompting on graphs.
\end{IEEEbiographynophoto}

\begin{IEEEbiographynophoto}{Zhenghao Liu}
is currently a Msc student with the School of Computer Science and Technology, University of Science and Technology of China. He received his B.S.~Degree in computer science from Shanghai University of Finance and Economics Shanghai, China in 2020.  His current research focuses on graph-based machine learning, graph mining, and prompting on graphs.
\end{IEEEbiographynophoto}

\begin{IEEEbiographynophoto}{Yuan Fang}
(Senior Member, IEEE) received the bachelor’s degree in computer science from the National University of Singapore in 2009 and the PhD degree in computer science from the University of Illinois at Urbana-Champaign in 2014. He is currently an assistant professor with the School of Computing and Information Systems, Singapore Management University. His current research focuses on graph-based data mining and machine learning, and their applications.
\end{IEEEbiographynophoto}

\begin{IEEEbiographynophoto}{Zemin Liu}
is currently a senior research fellow with the School of Computing, National University of Singapore. He received his Ph.D. degree in Computer Science from Zhejiang University, Hangzhou, China in 2018, and B.S. Degree in Software Engineering from Shandong University, Jinan, China in 2012. His research interests lie in graph embedding, graph neural networks, and learning on heterogeneous information networks.
\end{IEEEbiographynophoto}

\begin{IEEEbiographynophoto}{Sihong Chen}
is currently a senior researcher at Tencent AI. Her current research focuses on computer vision and multi-modal machine learning.
\end{IEEEbiographynophoto}

\begin{IEEEbiographynophoto}{Xinming Zhang}
(Senior Member, IEEE) received the BE and ME degrees in electrical engineering from the China University of Mining and Technology, Xuzhou, China in 1985 and 1988, respectively, and the PhD degree in computer science and technology from the University of Science and Technology of China, Hefei, China in 2001. Since 2002, he has been with the faculty of the University of Science and Technology of China, where he is currently a professor with the School of Computer Science and Technology. From 2005 to 2006, he was a visiting professor with the Department of Electrical Engineering and Computer Science, Korea Advanced Institute of Science and Technology, Daejeon, Korea. His research interest includes wireless networks, Big Data and deep learning. He has published more than 100 papers. He won the second prize of Science and Technology Award of Anhui Province of China in Natural Sciences in 2017. He won the awards of Top Reviewers (1\%) in Computer Science \& Cross Field by Publons in 2019.
\end{IEEEbiographynophoto}
\renewcommand\thesubsection{\Alph{subsection}}
\renewcommand\thesubsubsection{\thesubsection.\arabic{subsection}}

\end{document}